\documentclass{article}

\usepackage{PRIMEarxiv}

\usepackage[utf8]{inputenc} 
\usepackage[T1]{fontenc}    
\usepackage{hyperref}       
\usepackage{url}            
\usepackage{booktabs}       
\usepackage{amsfonts}       
\usepackage{nicefrac}       
\usepackage{microtype}      
\usepackage{lipsum}
\usepackage{fancyhdr}       
\usepackage{graphicx}       
\graphicspath{{media/}}     
\usepackage{amsmath}
\usepackage{cleveref}

\title{PoseFM: Relative Camera Pose Estimation Through Flow Matching
}

\author{
  Dominik Kuczkowski \& Laura Ruotsalainen \\
  Department of Computer Science \\
  University of Helsinki \\
  Finland \\
  \texttt{\{name.surname\}@helsinki.fi}\\
}

\begin{document}
\maketitle

\begin{abstract}
Monocular visual odometry (VO) is a fundamental computer vision problem with applications in autonomous navigation, augmented reality and more. While deep learning-based methods have recently shown superior accuracy compared to traditional geometric pipelines, particularly in environments where handcrafted features struggle due to poor structure or lighting conditions, most rely on deterministic regression, which lacks the uncertainty awareness required for robust applications. We propose PoseFM, the first framework to reformulate monocular frame-to-frame VO as a generative task using Flow Matching (FM). By leveraging FM, we model camera motion as a distribution rather than a point estimate, learning to transform noise into realistic pose predictions via continuous-time ODEs. This approach provides a principled mechanism for uncertainty estimation and enables robust motion inference under challenging visual conditions. In our evaluations, PoseFM achieves strong performance on TartanAir, KITTI and TUM-RGBD benchmarks, achieving the lowest absolute trajectory error (ATE) on some of the trajectories and overall being competitive with the best frame-to-frame monocular VO methods. Code and model checkpoints will be made available at https://github.com/helsinki-sda-group/posefm.
  
\end{abstract}


\section{Introduction}
\label{sec:intro}
\begin{figure}[tb]
  \centering
  \includegraphics[width=\textwidth]{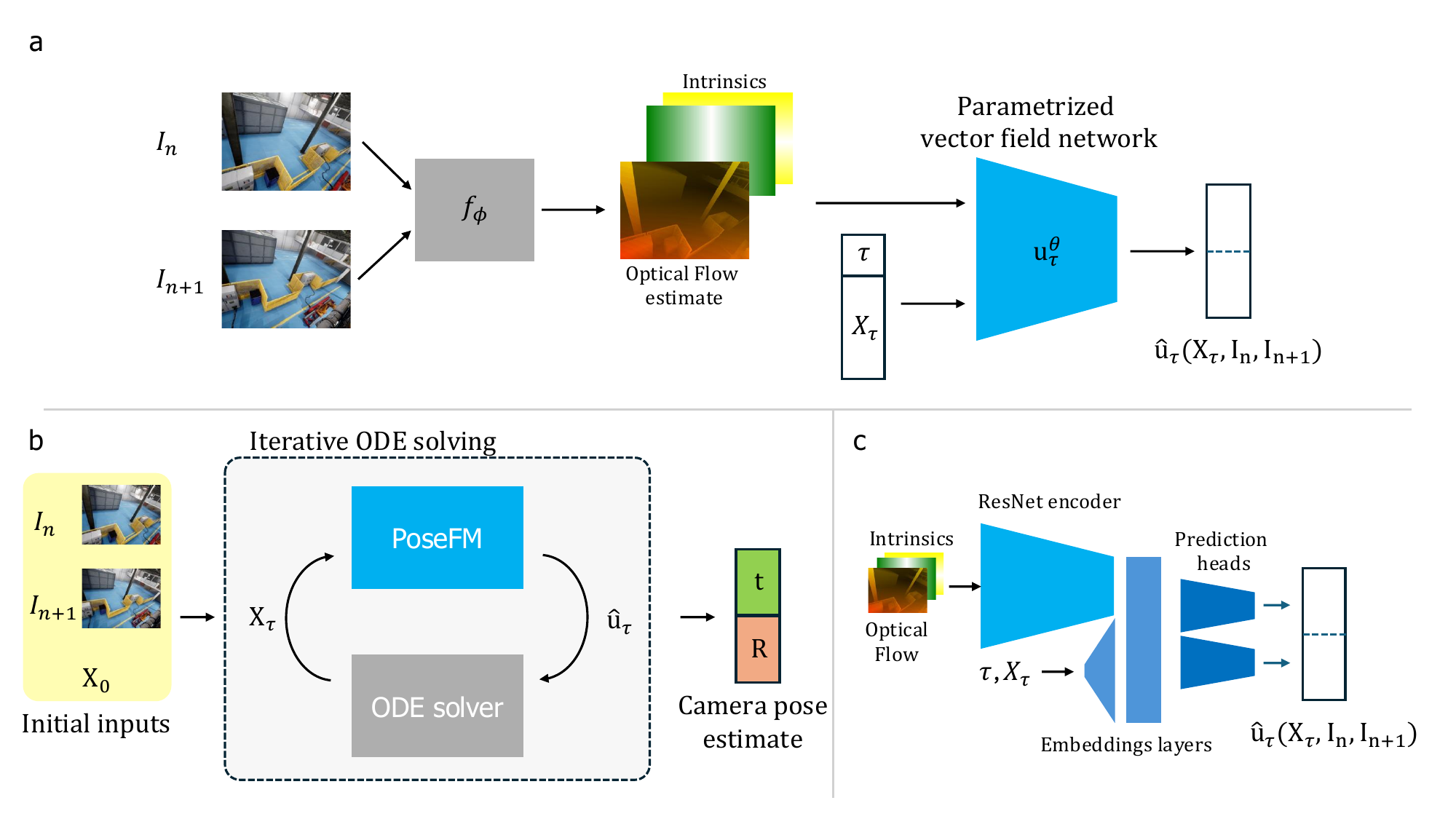}
  \caption{Overview of the PoseFM framework. \textbf{(a) PoseFM Pipeline}: The pipeline consists of an optical flow estimator $f_\phi$ and parametrized vector field network. The output of the pipeline is a point estimate of the vector field $\hat{u}_\tau$. \textbf{(b) Inference Procedure}: Given an image pair $(I_t, I_{t+1})$, we sample a pose $X_0$ from the initial distribution and numerically integrate the learned vector field using an ODE solver to recover the final camera motion. \textbf{(c) Model Architecture}: A detailed view of the network parameterizing $u_\tau$.}
  \label{fig:posefm}
\end{figure}

Monocular visual odometry (VO) is a fundamental computer vision (CV) problem, with many applications in robotics, autonomous navigation, and augmented reality. Despite years of research, VO still remains challenging , due to issues such as textureless surfaces, motion blur, dynamics scenes, or changing illumination.

Existing VO methods can be broadly divided into geometry-based and learning-based. Geometry-based methods \cite{orbslam3, dso} rely on explicit geometric constraints, such as epipolar geometry and photometric consistency, and typically offer strong interpretability and generalization under well-modeled assumptions. However, they often degrade in performance when these assumptions are violated. Learning-based methods \cite{dpvo, wang2021tartanvo, franccani2025tsformervo, Wang2024mambavo, li2026cuvo} on the other hand, leverage deep neural networks to learn motion estimation directly from data. They often use geometric or photometric constraints implicitly or as training objectives, offering improved robustness in challenging scenarios at the cost of reduced interpretability and, in some cases, limited generalization beyond the training distribution. While optical flow based methods have achieved strong performance in visual odometry, they are particularly prone to failure in challenging scenarios, compared to feature-based approaches \cite{pajula2023novel}. However, feature-based approaches use sparse correspondences while optical flow provides dense pixel-level motion cues that encode local geometric constraints between frames and therefore integrates naturally to deep learning frameworks. 

Recently, generative methods—particularly diffusion models \cite{ho2020denoising} and flow matching (FM) \cite{lipman2023flow}—have demonstrated remarkable success across a wide range of computer vision problems. In domains such as image synthesis, depth estimation, and structure-from-motion, diffusion-based approaches have been shown to effectively model complex, high-dimensional distributions and estimate global scene properties by iteratively refining noisy predictions \cite{rombach2022ldm}, \cite{gui2025depthfm}, \cite{wang2023posediffusion}.

Despite this progress, the application of diffusion-based or flow-based generative models to VO remains largely unexplored. To the best of our knowledge, prior work has not investigated diffusion models or FM formulations for estimating camera motion in a VO setting. This gap is notable because VO is inherently affected by ambiguity, perceptual aliasing, and accumulated uncertainty, particularly in challenging visual conditions. In such settings, modeling a distribution over possible motions, rather than committing to a single deterministic estimate via regression, offers clear advantages.

Therefore, in this work, we explore a FM-based generative perspective on VO. FM provides an alternative to diffusion modeling that learns continuous probability flows between simple base distributions and complex target distributions, enabling efficient training and sampling. By framing VO as a generative modeling problem over camera motions, we aim to leverage the strengths of flow-based methods—such as stability, flexibility, and principled uncertainty modeling—while addressing the unique challenges of motion estimation from visual data. 

Our contributions are as follows:
\begin{itemize}
    \item The PoseFM Framework: We introduce the first formulation of monocular VO as a generative task using Flow Matching. Our method integrates optical flow to provide dense visual guidance and employs flow matching to overcome the aforementioned challenges by inferring the global camera motion.
    \item We provide an evaluation of the generative modeling based approach for VO and show its competitiveness with regression-based methods, while providing inherent uncertainty estimation.
\end{itemize}

\section{Related Work}

\paragraph{Learning-based Visual Odometry} Traditional monocular VO approaches rely on explicit geometric constraints and estimate camera motion through nonlinear optimization. \cite{orbslam3, dso}. While the traditional approaches offer good generalizability, they often fail in challenging scenarios with dynamic cameras creating motion blur, moving objects, low lighting  or feature-sparsity. Learning-based approaches try to solve these issues, by improving robustness of monocular VO. Among them, we can differentiate between methods that rely on learned global optimization \cite{dpvo, Wang2024mambavo, Teed2021droidslam} and frame-to-frame methods, that directly regress a camera pose between frames \cite{wang2021tartanvo, li2026cuvo, shen2022dytanvo}. 

The methods based on learned optimization offer improved global consistency of estimated trajectories, but at the cost of relying on more complex pipelines, which makes them harder to train. 

Among the frame-to-frame methods, TartanVO \cite{wang2021tartanvo} was the first to propose a learning-based model generalizable to many datasets. DytanVO \cite{shen2022dytanvo} extended TartanVO with motion segmentation to improve performance in dynamic scenes. CUVO \cite{li2026cuvo} was built on a similar foundation as the two previous methods, predicting the camera pose based on an optical flow estimate. The novelty in CUVO, was the introduction of Context Attention and uncertainty awareness in the pose predictor. Our method falls into the frame-to-frame category with optical flow as the basis of camera pose estimation.

\paragraph{Generative Models in Computer Vision}

Generative models, particularly diffusion models \cite{ho2020ddpm, song2021scorebased} and flow matching based methods \cite{lipman2023flow}, have recently become a widely adopted paradigm in computer vision. Both approaches learn to transform samples from a simple base distribution into samples from a complex data distribution through an iterative refinement process. Their rise in popularity is largely driven by remarkable success in image generation \cite{ho2020ddpm, rombach2021latentdiffusion}, but they have since been extended to a variety of geometry-related tasks.

While diffusion-based methods such as PoseDiffusion \cite{wang2023posediffusion} and RayDiffusion \cite{zhang2024raydiffusion} demonstrate strong performance for global pose estimation in Structure-from-Motion (SfM), these methods operate in multi-view, batch setting where iterative sampling is advantageous. In contrast, VO is a frame-to-frame problem that requires efficient and temporally consistent motion estimation under real-time constraints.

FM learns deterministic transport between distributions via continuous-time velocity fields and  provides a computationally efficient and geometrically aligned alternative. FM has also been successfully applied to geometric prediction problems, including depth estimation \cite{gui2025depthfm} and 3D pose estimation \cite{wang2026fmpose3d}, where improved training stability and faster sampling compared to diffusion-based methods are reported. 

To the best of our knowledge, FM has not yet been explored for visual odometry, particularly in a frame-to-frame setting.

\paragraph{Uncertainty Estimation in Learning-based Visual Odometry}
Uncertainty estimation is a crucial component of VO, both for robust motion estimation and for its use in safety-critical downstream tasks. In learning-based VO, uncertainty is typically modeled at the pixel or feature level. For example, D3VO \cite{Yang_2020_CVPR} models photometric uncertainty at the pixel level to improve robustness in learned depth estimation, which is then used within a direct VO pipeline. Similarly, depth uncertainty estimation methods \cite{Wang_uncertainty} model photometric sensitivity to better characterize unreliable measurements. 

In fully deep learning based VO, CoProU-VO \cite{Keuper2026} improves robustness in unsupervised monocular VO by propagating photometric uncertainty across frames to better identify regions violating the static scene assumption, while CUVO \cite{li2026cuvo} filters unreliable regions and incorporates an uncertainty selection module to mitigate the impact of dynamic objects and noise. These approaches effectively down-weight ambiguous or corrupted measurements, leading to improved local robustness.

However, such uncertainty formulations remain predominantly local and unimodal, focusing on measurement noise rather than modeling distributions over camera motion itself. In scenarios with limited parallax, perceptual aliasing, or motion blur, multiple camera motions may explain the observations equally well even when photometric consistency is satisfied. In these cases, pixel-level uncertainty weighting alone cannot represent multi-modal pose hypotheses, and the system typically collapses to a single deterministic estimate.

In contrast, we model camera motion as a distribution over poses, enabling ambiguity to be retained rather than prematurely resolved at each frame pair. This generative formulation allows uncertainty to be expressed directly in pose space, addressing failure modes arising from inherent geometric ambiguity rather than solely from noisy measurements.

\section{Method}
We propose PoseFM, a novel FM-based VO method that learns a distribution of camera motions. To perform this task, PoseFM employs a deep learning end-to-end pipeline. The pipeline consists of an optical flow estimator and FM-based camera motion estimator. It is inspired by classical computer vision architectures and learning-based frame-to-frame VO methods \cite{wang2021tartanvo, shen2022dytanvo, li2026cuvo}. Our FM-based architecture is inspired by TartanVO, allowing for a controlled comparison between generative modeling and deterministic regression. 

\subsection{Generative Formulation of VO}
The goal of VO is to estimate the rigid-body motion $T\in SE(3)$ between two consecutive image frames $I_n,I_{n+1}$. The transformation $T$ consists of a rotation $R\in SO(3)$, the group of 3D rotations, and a translation $t\in R^3$. Since $SE(3)$ is a nonlinear Lie group, motion estimation is commonly performed in its associated Lie algebra $\mathfrak{se}(3)$, which provides a locally linear, minimal representation. 

We formulate the estimation of $T$ as a generative problem. Let $p(T|I_n,I_{n+1})$ be a distribution of camera motions conditioned on the visual input. This distribution is unknown in closed form. Our goal is to sample from it by employing FM.

The main idea behind FM is to estimate a parametrized time-dependent vector field $u_\tau^{\theta}$, which points in the direction of a target distribution $p_1$, given a sample from an initial distribution $p_0$ \cite{lipman2023flow}. $p_0$ should be easy to sample, so that during inference we can draw samples from it and then use $u_\tau^{\theta}$ to transform the samples, in such a way that they are distributed according to $p_1$. In our case, we utilize separate initial distributions for the translation $t$ and rotation $R$ components of the motion $T$. We choose isotropic normal for $t$ \cref{eq:t_dist} and a uniform distribution on $\mathrm{SO}(3)$ group for $R$ \cref{eq:R_dist}. Our target distribution is the distribution of relative camera poses $p_1=p(T|I_n,I_{n+1})$.

\begin{equation}
    \label{eq:t_dist}
    t\sim \mathcal{N}(0, I), \text{where}~ t\in \mathbb{R}^3 
\end{equation}
\begin{equation}
    \label{eq:R_dist}
    R\sim \mathcal{U}(\mathrm{SO}(3)), \text{where}~R\in \mathrm{SO}(3)
\end{equation}

We represent the rotation $R$ in the Lie algebra $\mathfrak{so}(3)$. This allows us to apply flow matching in Euclidean space, avoiding the need to learn vector fields directly on the rotation manifold.

\subsection{Training Objective and Guidance}
Our method is trained using the Conditional Flow Matching (CFM) loss with an Optimal Transport path \cite{lipman2023flow}. 
Let \(u_{\tau}^\theta:\mathbb{R}^d\times[0,1]\to\mathbb{R}^d\) denote the parametrized time-dependent vector field, where $d$ is the dimension of the estimation space. We use $\tau$ as the symbol of time to differentiate it from the translation $t$.
The CFM loss is defined as:
\begin{equation}
    \mathcal{L}(\theta)= \mathbb{E}_{\tau, X_0, X_1}\big\lVert u_{\tau}^\theta(X_{\tau}) - (X_1 - X_0)\big\rVert_2^2,   
\end{equation}
where $\tau\sim\mathcal{U}[0,1]$, $X_0 \sim p_0$, $X_1 \sim p_1$, and $X_{\tau} = (1-\tau)X_0 + \tau X_1$. $X_0, X_1$ are samples from the initial and target distributions, respectively.

We want to estimate the conditional distribution $p(T|I_n,I_{n+1})$, therefore we additionally condition the vector field. In the FM literature this is referred to as guidance. Our vector field with guidance is:
\begin{equation}
    u_{\tau}^\theta:\mathbb{R}^d\times[0,1]\times\mathbb{R}^{c\times h\times w}\to\mathbb{R}^d,
\end{equation}
where $d$ is the dimension of the camera motion space, $c$ is the number of channels of the visual guidance signal, and ($h$, $w$) determine the frame size of the visual guidance signal. Our training objective with guidance is defined as:
\begin{equation}
    \mathcal{L}(\theta)= \mathbb{E}_{\tau, X_0, X_1, I}\big\lVert u_{\tau}^\theta(X_{\tau}, f_\phi(I_n, I_{n+1})) - (X_1 - X_0)\big\rVert_2^2,   
\end{equation}
where $f_\phi$ is a frozen encoder of the visual input.

\subsection{End-To-End Pipeline}
Our method follows a two-stage architecture illustrated in \cref{fig:posefm}a. The first stage functions as an encoder $f_\phi$ that produces a visual guidance signal for the time-dependent vector field $u_\tau^\theta$. Following common practice in frame-to-frame VO \cite{shen2022dytanvo, li2026cuvo} we employ an optical flow estimator as the front-end. We evaluate two variants of front-ends, first one based on PWCNet \cite{Sun2017pwcnet} and second based on the state-of-the-art WAFT \cite{wang2026waft} optical flow estimator. In both cases we keep the parameters of the model frozen during training. For the PWCNet front-end we use weights from \cite{wang2021tartanvo}.

The second stage parametrizes the time-dependent vector field $u_\tau^\theta$. Its architecture extends the TartanVO\cite{wang2021tartanvo} design by introducing modifications required to support FM, visible in \cref{fig:posefm}c. The model is based on a ResNet \cite{resnet} feature extractor, followed by two separate prediction heads for translation and rotation. We add additional embedding layers for the FM-specific inputs, namely the time variable $\tau$ and the intermediate pose state $X_{\tau}$. These embeddings are fused with the visual features between the ResNet encoder and the prediction heads.

\subsection{Inference-Time Camera Pose Estimation}
The parametrized vector field $u_{\tau}^\theta$ only gives a direction in which the samples need to change, so in FM the actual sample generation at inference time is performed by integrating an ordinary differential equation (ODE).
The ODE formulation of FM is:
\begin{equation}
    \frac{dx}{d\tau}\psi(\tau, x) = u_\tau(\psi(\tau,x)),\qquad x(0)\sim p_0,
\end{equation}

where \(\psi:[0,1]\times\mathbb{R}^d\to\mathbb{R}^d\) is a time-dependent flow determined by the vector field $u_\tau$. The ODE is solved by integrating the parametrized vector field $u_\tau^\theta$. We employ an ODE solver to numerically integrate the vector field and obtain final motion estimate.

\subsection{Uncertainty Quantification}
The inherent advantage of using FM-based formulation to deterministic regression-based models is that it naturally enables uncertainty assessment. During inference the ODE solver is initialized from a randomly sampled initial camera pose. This creates an opportunity to sample multiple initial conditions and evaluate the properties of samples obtained from the ODE solver. Assuming a well-trained model, high variance of the obtained samples suggests low-confidence of the model, and vice versa. Unlike approaches that explicitly predict uncertainty parameters, this formulation provides uncertainty directly from the variability of generated samples and reduces the risk of the model becoming overconfident in incorrect predictions. 

\section{Experiments}
In this section we describe the implementation details of our method and the training specifics. We also include the evaluation results.

\subsection{Implementation Details}
\paragraph{Training}
Our method was trained on the challenging TartanAir dataset \cite{wang2020tartanair}, using a single Nvidia Tesla V100 GPU. Our model was implemented in PyTorch. We followed separate training procedures for the two different variants of our method.

For PoseFM+PWC we first trained the vector field $u_\tau^\theta$ on the ground truth (GT) optical flow (OF) data. We performed 100 epochs with batch size 128 and Adam optimizer. The initial learning rate (LR) was set to 1e-4. Following a step decay schedule, the LR was reduced by a factor of 0.5 after 50 epochs. We then fine-tuned the vector field network on OF predictions from PWCNet. We performed 50 epochs with batch size 96 and LR 5e-4. The OF encoder was frozen during the whole training.

For PoseFM+WAFT, we first used WAFT to precompute OF for the training dataset. We then trained the vector field network on the precomputed OF for 100 epochs using the same optimizer and learning rate schedule as in the previous setting.

Following \cite{wang2021tartanvo, li2026cuvo}  we used an additional intrinsics layer to enable generalization, in both setups. We also applied the RandomCropResize augmentation as in \cite{wang2021tartanvo}.

\paragraph{Inference}
To obtain estimates at inference time, we first sampled a pose $X_0$ from the initial distribution and passed it to the PoseFM pipeline, along with images $I_n$, $I_{n+1}$. We numerically integrated the learned vector field $\hat{u}_\tau$ from $\tau=0$ to $\tau=1$ using an ODE solver. The number of integration steps was a hyperparameter, which we set to 5. Our ODE solver used the midpoint method. As a result of the integration we obtained $X_1$, which was a single motion estimate. The whole process is shown in \cref{fig:posefm}b.

Following other frame-to-frame VO methods \cite{wang2021tartanvo, shen2022dytanvo, li2026cuvo} we treat our translation estimate as up-to-scale. Therefore, during inference we perform translation scale alignment as in \cite{wang2021tartanvo, zhou2017unsdepth, yin2018geonet, li2026cuvo}.

\paragraph{Uncertainty Estimation}
Our FM formulation requires sampling an initial pose $X_0$ to initialize the ODE solver. 
This stochastic initialization naturally induces a distribution over predicted camera motions. 

At test time, we draw $m$ independent samples $\{X_0^{(i)}\}_{i=1}^m$ from the initial distribution and solve the ODE for each initialization, obtaining a set of pose predictions $\{X_1^{(i)}\}_{i=1}^m$. 
The empirical variance of these predictions provides an estimate of model uncertainty in pose space. 
In practice, we compute the sample mean and standard deviation across the $m$ predicted poses, which serve as confidence measures for the estimated motion. In our experiments we set $m=10$ and take the mean value as final motion estimate.

\subsection{Results on TartanAir}
We evaluate our method on the challenging TartanAir test set trajectories. Similarly to other works \cite{wang2021tartanvo, dpvo, li2026cuvo, Wang2024mambavo}, we measure the Absolute Trajectory Error (ATE). The results are reported after a 7-DoF alignment that fixes the scale ambiguity problems of monocular VO. We compare our method to other frame-to-frame VO methods, including TartanVO \cite{wang2021tartanvo} and DytanVO \cite{shen2022dytanvo}—with results as reported in \cite{li2026cuvo}—as well as CUVO \cite{li2026cuvo} itself. Additionally we include the results for optimization based multi-frame methods, including the classical ORB-SLAM3 \cite{orbslam3} and learning-based DPVO \cite{dpvo} and MambaVO \cite{Wang2024mambavo}.
\begin{table}[ht]
\centering
\caption{ATE [m] comparison on TartanAir test set. \textbf{Bold} indicates the best overall method, while \underline{underline}
denotes the best frame-to-frame method. Methods marked with $^\dagger$ use multiple frames for pose optimization.}
\label{tab:tartan_comparison}
\scriptsize 
\setlength{\tabcolsep}{1.0pt} 
\resizebox{\textwidth}{!}{
    \begin{tabular}{lcccccccc|cccccccc|c}
    \toprule
    & ME & ME & ME & ME & ME & ME & ME & ME & MH & MH & MH & MH & MH & MH & MH & MH \\
     & 000 & 001 & 002 & 003 & 004 & 005 & 006 & 007 & 000 & 001 & 002 & 003 & 004 & 005 & 006 & 007 & Avg\\ \midrule
    
    ORB-SLAM3 \cite{orbslam3}$^\dagger$ & 13.61 & 16.86 & 20.57 & 16.00 & 22.27 & 9.28 & 21.61 & 7.74 & 15.44 & 2.92 & 13.51 & 8.18 & 2.59 & 21.91 & 11.70 & 25.88 & 14.38\\ 
    DPVO \cite{dpvo}$^\dagger$ & \textbf{0.16} & \textbf{0.11} & \textbf{0.11} & \textbf{0.66} & \textbf{0.31} & \textbf{0.14} & \textbf{0.30} & \textbf{0.13} & \textbf{0.21} & 0.04 & 0.04 & 0.08 & 0.58 & \textbf{0.17} & \textbf{0.11} & 0.15 & 0.21\\
    MambaVO \cite{Wang2024mambavo}$^\dagger$ & - & - & - & - & - & - & - & - & 0.24 & \textbf{0.02} & \textbf{0.03} & \textbf{0.02} & 0.46 & 0.18 & 0.13 & \textbf{0.05} & -\\
    \midrule
    TartanVO \cite{wang2021tartanvo}& 27.3 & 0.86 & \underline{0.64} & 7.18 & 2.02 & \underline{0.58} & 4.12 & \underline{0.42} & 2.12 & 0.31 & 1.28 & 1.09 & 0.99 & 1.4 & 1.74 & \underline{1.42} & 3.34\\ 
    DytanVO \cite{shen2022dytanvo} & 25.95 & 1.36 & 1.17 & 6.94 & 2.75 & 0.96 & 4.46 & 0.89 & 5.10 & \underline{0.22} & 1.62 & \underline{0.79} & 1.29 & 4.46 & 2.06 & 2.36 & 3.90\\
    CUVO \cite{li2026cuvo} & 12.94 & \underline{0.52} & 2.00 & \underline{4.55} & \underline{1.48} & 0.59 & \underline{0.63} & 0.65 & 2.95 & 0.62 & \underline{0.79} & 0.84 & \underline{\textbf{0.40}} & \underline{0.69} & \underline{0.75} & 1.46 & 1.99 \\
    PoseFM+PWC (Ours) & \underline{11.66} & 1.07 & 2.85 & 9.81 & 2.29 & 0.87 & 5.07 & 1.12 & 3.46 & 0.28 & 1.98 & 1.13 & 0.95 & 2.55 & 2.65 & 1.53 & 3.08 \\
    PoseFM+WAFT (Ours) & 12.88 & 1.36 & 1.35 & 7.64 & 3.94 & 0.76 & 4.55 & 1.06 & \underline{1.74} & 0.42 & 1.5 & 0.82	& 0.66 & 2.45 & 4.08 & 2.07 & 2.96 \\
    \bottomrule
    \end{tabular}
}
\end{table}

In \cref{tab:tartan_comparison}, we summarize the results on the TartanAir test set. Among frame-to-frame approaches, both variants of PoseFM achieve competitive performance. In particular, PoseFM+PWC obtains an average ATE of 3.08, improving over TartanVO (3.34) by approximately 8\%. Moreover, each PoseFM variant achieves the best performance on at least one trajectory, demonstrating robustness across different motion patterns and scene conditions.

As expected, optimization-based multi-frame methods such as DPVO and MambaVO achieve the lowest overall ATE due to their ability to perform temporal refinement and global trajectory optimization. However, our method achieves competitive performance despite operating in a purely frame-to-frame setting without explicit multi-frame optimization. This result highlights the effectiveness of the proposed FM-based formulation, demonstrating that modeling camera motion as a distribution enables accurate motion estimation even without global trajectory refinement, while achieving comparable or improved accuracy relative to  prior pairwise approaches.

\subsection{Results on KITTI}
We further evaluate our method on KITTI odometry sequences 00–10 \cite{geiger2012kitti}. KITTI is a standard benchmark in VO, which we utilize to test generalization capabilities of our model. We report the ATE metric values, without any fine-tuning on the KITTI dataset. We compare ourselves to the same set of baselines as in the case of TartanAir.
\begin{table}[ht]
\centering
\caption{ATE [m] comparison on KITTI odometry sequences 0-10. \textbf{Bold} indicates the best VO method, while \underline{underline} denotes second best VO method. Methods marked with $\dagger$ use multiple frames for pose optimization.}
\label{tab:kitti_results}
\scriptsize 
\setlength{\tabcolsep}{2.0pt} 
\resizebox{\textwidth}{!}{
    \begin{tabular}{lccccccccccc|c} 
    \toprule
    \textbf{Sequence} & \textbf{00} & \textbf{01} & \textbf{02} & \textbf{03} & \textbf{04} & \textbf{05} & \textbf{06} & \textbf{07} & \textbf{08} & \textbf{09} & \textbf{10} & \textbf{Avg}\\ \midrule

    ORB-SLAM3 \cite{orbslam3}$^\dagger$ & 6.77 & \texttimes & 30.50 & 1.04 & 0.93 & 5.54 & 16.61 & 9.70 & 60.69 & 7.89 & 8.65 & -\\ 
    DPVO \cite{dpvo}$^\dagger$ & 113.21 & \underline{12.69} & 123.40 & \underline{2.09} & \underline{0.68} & 58.96 & 54.78 & 19.26 & 115.90 & 75.10 & 13.63 & 53.03\\
    MambaVO \cite{Wang2024mambavo}$^\dagger$ & 112.39 & \textbf{8.16} & 93.78 & \textbf{1.80} & \textbf{0.66} & 56.51 & 57.19 & 17.90 & 116.01 & 73.56 & 14.37 & 50.21\\
    \midrule
    
    TartanVO \cite{wang2021tartanvo} & 69.11 & 53.19 & 78.78 & 2.70 & 1.99 & 55.18 & \textbf{10.50} & 13.87 & 48.16 & \underline{27.93} & 11.90 & 33.94\\ 
    DytanVO \cite{shen2022dytanvo}   & \textbf{43.14} & 30.83 & 64.94 & 4.36 & 1.05 & \underline{33.83} & 21.85 & 23.51 & 30.43 & 28.87 & \underline{10.50} & 26.66\\ 
    CUVO \cite{li2026cuvo} & \underline{44.73} & 69.92 & \textbf{52.41} & 2.82 & 1.16 & \textbf{23.90} & \underline{17.16} & \underline{6.07} & \textbf{24.22} & \textbf{14.38} & \textbf{4.39} & 23.74\\
    PoseFM+PWC (Ours) & 59.66 & 123.48 & \underline{63.36} & 6.42 & 3.55 & 49.74 & 32.42 & \textbf{4.56} & \underline{24.86} & 45.36 & 12.76 & 38.74 \\
    PoseFM+WAFT (Ours) & 175.68 & 44.4 & 276.85 & 17.02 & 13.9 & 101.19 & 54.62 & 41.38 & 218.05 & 68.17 & 59.75 & 97.36 \\
    \bottomrule
    \end{tabular}
}
\end{table}
As shown in \cref{tab:kitti_results}, PoseFM+PWC demonstrates competitive performance across sequences among the VO methods. It achieves the best result on one sequence and ranks second on two additional sequences, indicating strong robustness under diverse motion patterns. Notably, the performance gap between frame-to-frame and multi-frame learning-based methods is smaller on KITTI compared to TartanAir, and multi-frame approaches do not consistently outperform pairwise methods.

The PoseFM+WAFT variant underperforms relative to other methods on KITTI. We attribute this behavior to domain shift effects and sensitivity to the differing camera intrinsics and motion statistics of KITTI.

\subsection{Results on TUM-RGBD}
As a final benchmark we test our method on TUM-RGBD dataset \cite{sturm12tumrgbd}. This evaluation showcases the generalization capabilities of our method to indoor environments. We decided to only test the PoseFM+PWC variant as it showed better generalization capability on KITTI.
\begin{table}[ht]
\centering
\caption{ATE [m] comparison on TUM-RGBD. \textbf{Bold} indicates the best VO method, while \underline{underline} denotes the best frame-to-frame method. Methods marked with $\dagger$ use multiple frames for pose optimization.}
\label{tab:tum_results}
\scriptsize 
\setlength{\tabcolsep}{2.0pt} 
\resizebox{\textwidth}{!}{
    \begin{tabular}{lcccccccccc} 
    \toprule
    & \textbf{360} & \textbf{desk} & \textbf{desk2} & \textbf{floor} & \textbf{plant} & \textbf{room} & \textbf{rpy} & \textbf{teddy} & \textbf{xyz} & Avg. \\ \midrule

    ORB-SLAM3 \cite{orbslam3}$^\dagger$ & $\times$ & 0.016 & $\times$ & $\times$ & 0.038 & $\times$ & $\times$ & $\times$ & 0.005 & - \\ 
    DPVO \cite{dpvo}$^\dagger$ & 0.156 & 0.034 & 0.05 & 0.183 & 0.034 & 0.383 & 0.038 & 0.073 & \textbf{0.012} & 0.107  \\
    MambaVO \cite{Wang2024mambavo}$^\dagger$ & \textbf{0.108} & \textbf{0.021} & \textbf{0.037} & \textbf{0.034} & \textbf{0.022} & \textbf{0.372} & \textbf{0.031} & \textbf{0.048} & 0.013 & 0.076 \\
    \midrule
    
    TartanVO \cite{wang2021tartanvo} & 0.160 & 0.478 & 0.539 & 0.348 & 0.395 & 0.417 & \underline{0.05} & \underline{0.329} & 0.16 & 0.32 \\ 
    DytanVO \cite{shen2022dytanvo} & 0.188 & 0.159 & 0.224 & 0.191 & 0.343 & 0.53 & 0.053 & 0.508 & 0.131 & 0.259 \\
    CUVO \cite{li2026cuvo} & 0.206 & \underline{0.078} & 0.106 & 0.340 & \underline{0.207} & \underline{0.331} & 0.075 & 0.334 & \underline{0.061} & 0.193 \\
    PoseFM+PWC (Ours) & \underline{0.146} & 0.108 & \underline{0.085} & \underline{0.18} & 0.36 & 0.454 & 0.051 & 0.36 & 0.087 & 0.203 \\
    \bottomrule
    \end{tabular}
}
\end{table}

The results shown in \cref{tab:tum_results} demonstrate strong generalization to indoor trajectories. Our method achieves the best result among the frame-to-frame methods on 3 out of 9 scenes. It nearly matches the average ATE of the best to date frame-to-frame method, CUVO, being only 5\% worse, while reducing the average ATE for 36\% in comparison with TartanVO. 

\subsection{Ablations}
We conduct a series of ablation studies on the TartanAir test set to analyze the contribution of key design choices, including fine-tuning strategy, integration steps during inference, and the choice of optical flow front-end.

\paragraph{Role of Fine-Tuning on Predicted Flow}

To evaluate the impact of fine-tuning on predicted optical flow, we trained a model using only ground-truth (GT) flow supervision and evaluated it using PWCNet and WAFT optical flow predictions. Without fine-tuning, PoseFM+PWC exhibits approximately 20\% higher average ATE compared to the fine-tuned variant. In this case the result highlights the importance of adapting the vector field to the distribution of predicted optical flow, which differs from GT flow due to estimation noise and artifacts. In contrast PoseFM fine-tuned with WAFT reaches higher 11\% higher ATE than the not fine-tuned baseline. Therefore, for PoseFM+WAFT we modify the training procedure to perform training directly on the OF estimates.

\paragraph{Number of Integration Steps}

We analyze the effect of the number of integration steps used during inference by evaluating the PoseFM+WAFT model with 2, 5 and 10 integration steps. The average ATE values are 3.03, 2.90 and 2.98 respectively. The best performance is obtained with 5 integration steps, although the differences across configurations are relatively small. This suggests that a moderate number of steps is sufficient to approximate the continuous flow dynamics, while further increasing the integration depth does not yield additional gains. Notably, even 2 integration steps provide competitive accuracy while offering improved computational efficiency.

\paragraph{Optical Flow Front-End}

To study the influence of the optical flow front-end, we trained the pose estimation model on GT flow and evaluated it using different predicted flow sources. Using PWCNet and WAFT results in similar average ATE values of 3.71 and 3.59, respectively. This indicates that the proposed flow-matching formulation is robust to variations in optical flow quality and can effectively adapt to different front-end estimators. The relatively small performance gap suggests that pose estimation accuracy is not solely determined by the flow backbone but also by the downstream motion modeling.

\section{Conclusion}
In this work, we introduced PoseFM, a novel flow-matching-based formulation for visual odometry that models camera motion through a learned continuous vector field. Unlike traditional regression-based approaches, our method adopts a generative perspective on frame-to-frame pose estimation, enabling improved robustness to noisy optical flow and facilitating uncertainty estimation through sampling.

Through experiments on TartanAir, KITTI and TUM-RGBD we demonstrated that PoseFM achieves competitive performance among frame-to-frame visual odometry methods, attaining the best results on multiple challenging sequences. In direct comparison with a regression based TartanVO, our method achieves better performance on average, showcasing the applicability of flow matching to visual odometry.

We believe that PoseFM is a promising framework for visual odometry.
Future work includes extending it to multi-frame optimization, incorporating more powerful network architectures, and building an image to pose pipeline, without reliance on optical flow.

\section*{Acknowledgments}
The authors acknowledge the research environment provided by ELLIS Institute Finland. This work was supported by the Research Council of Finland Flagship program Finnish Center for Artificial Intelligence FCAI, and the Department of Computer Science, University of Helsinki.

\bibliographystyle{unsrt}  
\bibliography{references.bib}

\end{document}